\pdfoutput=1

\documentclass[11pt]{article}

\usepackage{ACL2023}

\usepackage{times}
\usepackage{latexsym}

\usepackage[T1]{fontenc}

\usepackage[utf8]{inputenc}

\usepackage{microtype}

\usepackage{inconsolata}

\usepackage{algorithm}
\usepackage{algorithmic}
\usepackage{amsthm}
\usepackage{amsmath}
\usepackage{amssymb}
\usepackage{CJKutf8}
\usepackage{mathrsfs}
\usepackage{bm}
\usepackage{bbding}
\usepackage{color}
\usepackage{graphicx}
\usepackage{makecell}
\usepackage{multirow}
\usepackage{microtype}
\usepackage{subfigure}

\title{Harvesting Event Schemas from Large Language Models}

\author{
\textbf{Jialong Tang}${}^{1,3}$, \textbf{Hongyu Lin}${}^{1}$, \textbf{Zhuoqun Li}${}^{1,3}$, \textbf{Yaojie Lu}${}^{1}$, \\
\textbf{Xianpei Han}${}^{1,2}$, \textbf{Le Sun}${}^{1,2,*}$\\
${}^{1}$Chinese Information Processing Laboratory ~ ${}^{2}$State Key Laboratory of Computer Science\\
Institute of Software, Chinese Academy of Sciences, Beijing, China\\
${}^{3}$University of Chinese Academy of Sciences, Beijing, China\\
\texttt{\{jialong2019,xianpei,sunle\}@iscas.ac.cn} \\
}

\begin{document}
\maketitle
\begin{abstract}
Event schema provides a conceptual, structural and formal language to represent events and model the world event knowledge.
Unfortunately, it is challenging to automatically induce high-quality and high-coverage event schemas due to the open nature of real-world events, the diversity of event expressions, and the sparsity of event knowledge. 
In this paper, we propose a new paradigm for event schema induction -- knowledge harvesting from large-scale pre-trained language models, which can effectively resolve the above challenges by discovering, conceptualizing and structuralizing event schemas from PLMs. 
And an \textbf{E}vent \textbf{S}chema \textbf{H}arvest\textbf{er} (\textbf{ESHer}) is designed to automatically induce high-quality event schemas via in-context generation-based conceptualization, confidence-aware schema structuralization and graph-based schema aggregation.
Empirical results show that ESHer can induce high-quality and high-coverage event schemas on varying domains.
\let\thefootnote\relax\footnotetext{${}^{*}$Corresponding authors.}
\footnote{Our source codes with corresponding experimental datasets will be openly available at \url{https://github.com/TangJiaLong/Event-Schema-Harvester}.
}
\end{abstract}

\section{Introduction}
\label{sec:introduction}
Event is one of the basic units for human beings to understand and experience the world~\citep{jackendoff-1992-semantic}. 
An event is a specific occurrence involving multiple participants, such as \emph{bombing}, \emph{election}, and \emph{marriage}.
To represent events and model the world event knowledge, event schema provides a conceptual, structural and formal language which can describe the types of events, the semantic roles (slots) of specific events, and the relations between different events.
Specifically, an event schema is a frame such as ``\emph{\textbf{Type}: bombing, \textbf{Slots}: perpetrator, victm, target, instrument}''~\citep{chambers-jurafsky-2011-template}, which is central in event extraction~\citep{lu-etal-2021-text2event,lu-etal-2022-unified}, event relationship understanding~\citep{irwin-etal-2011-narrative,li-etal-2020-connecting,li-etal-2021-future}, and event knowledge base construction~\citep{zhang-etal-2020-transomcs,he-etal-2022-acquiring}. 
Due to its importance, it is critical to automatically discover and construct large-scale, high-quality, and high-coverage event schemas, i.e., event schema induction.

\begin{figure}[!t]
\centering
\includegraphics[width=\columnwidth]{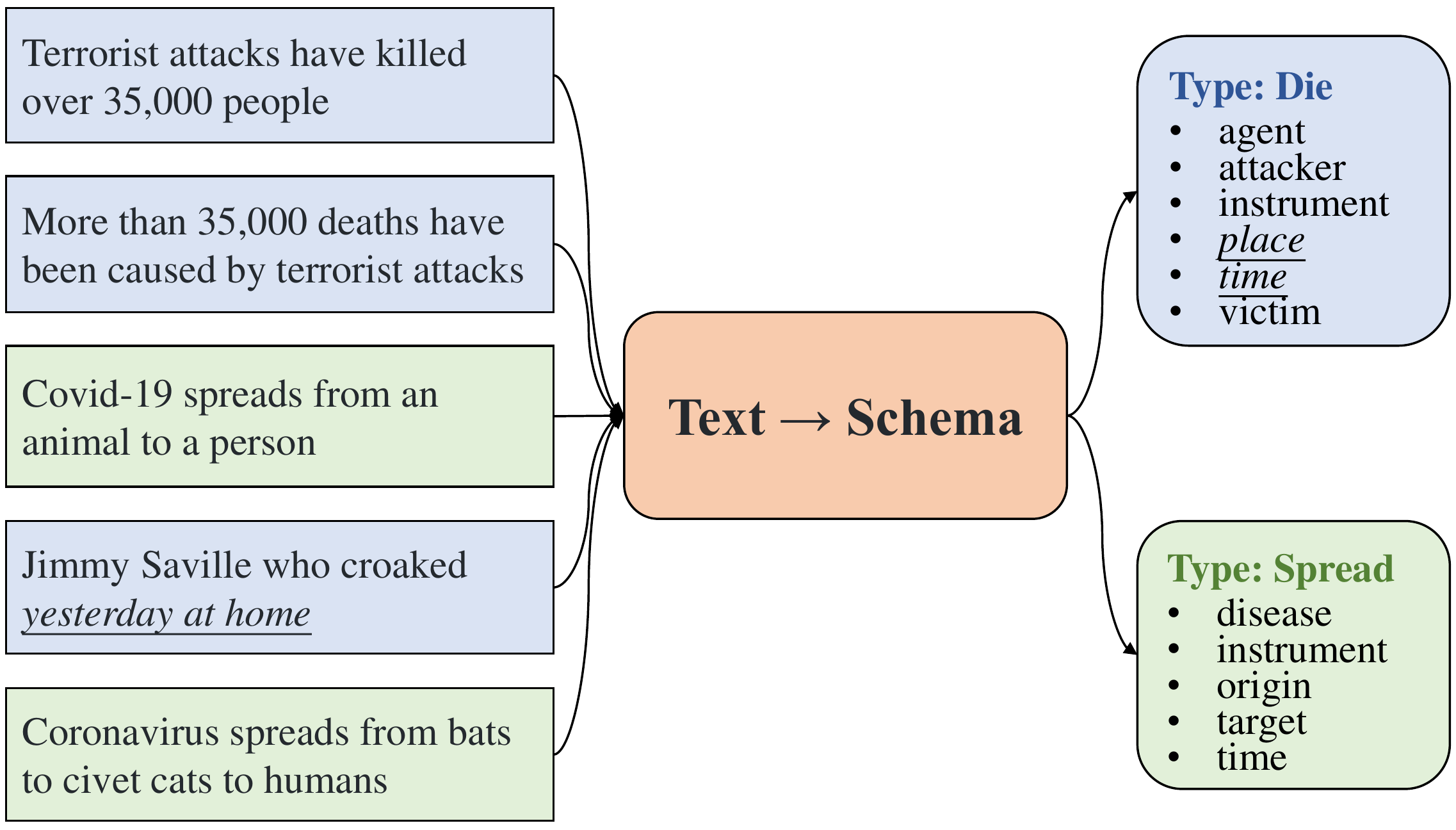}
\caption{
The event harvest paradigm for event schema induction, which induces high-quality event schemas from diverse event expressions and dispersed event knowledge on open domains, e.g., the top two texts on the left use different utterances to describe the same event while the fourth text complements the slots  ``\underline{\emph{place}}'' and ``\underline{\emph{time}}''.
}
\label{fig:introduction}
\end{figure}

Event schema induction, unfortunately, is a non-trivial task due to the open nature of real-world events, the diversity of event expressions, and the sparsity of event knowledge. 
Firstly, in real-world applications, the size of event types is very large and new types of events are constantly emerging. 
To address this open problem, event schemas must be induced automatically and with a high coverage on varying domains. 
Secondly, as shown in Figure~\ref{fig:introduction}, events are usually expressed using very different natural language utterances, therefore it is critical to normalize diverse event expressions by conceptualizing and structuralizing them into formal event schemas.
Finally, due to the economic principle of language~\citep{de-2011-course}, event expressions are mostly incomplete and many event arguments are missing. 
To resolve this sparsity problem, an event schema induction method must aggregate dispersed event knowledge across different expressions.

Up to recently, all most event schemas are still hand-engineered by human experts, which are expensive and labour-intensive (e.g., schemas in MUC~\citep{chinchor-etal-1993-evaluating}, ACE~\citep{doddington-etal-2004-automatic} and TAC-KBP~\citep{ji-grishman-2011-knowledge}). 
On the other hand, traditional automatic event schema induction methods still cannot overcome the open, diversity, and sparsity challenges.
For instance, bottom-up concept linking methods~\citep{huang-etal-2016-liberal,he-etal-2022-acquiring} discover event types/slots by parsing and linking event expressions to external schema resources such as FrameNet~\citep{baker-etal-1998-berkeley-framenet}, which are limited by the quality and the coverage of external schema resources.
Top-down clustering methods~\citep{chambers-2013-event,cheung-etal-2013-probabilistic,nguyen-etal-2015-generative,sha-etal-2016-joint,ahn-2017-inducing,yuan-etal-2018-open,liu-etal-2019-open,shen-etal-2021-corpus} cluster event expressions according to pre-defined schema templates (e.g., the 5W1H template, or templates with the predefined number of event types/slots), which are highly constrained by the pre-defined templates. 
To sum up, it remains a critical challenge to automatically discover schemas on open domains, normalise event schemas from diverse expressions, and aggregate dispersed knowledge from sparse descriptions.

In this paper, we propose a new paradigm for event schema induction -- knowledge harvesting from large pre-trained language models (PLMs), which can effectively address the open, diversity, and sparsity challenges.
The main idea is to automatically harvest open-domain and high-coverage event schemas from large PLMs and leverage the strong text generation and in-context learning abilities of PLMs for discovering, conceptualizing, and structuralizing event schemas.

Specifically, we design an \textbf{E}vent \textbf{S}chema \textbf{H}arvest\textbf{er} (\textbf{ESHer}), which automatically discovers and normalizes event types and their semantic roles via the following three components:
1) \emph{text conceptualization via in-context generation}, which can unsupervised-ly transform diverse event expressions into conceptualized event schema candidates based on in-context demonstrations;
2) \emph{confidence-aware schema structuralization}, which structuralizes event schemas by selecting and associating event types with their salient, reliable and consistent slots;
3) \emph{graph-based schema aggregation}, which aggregates dispersed event schemas via graph-based clustering.
In this way, the open, diversity, and sparsity challenges can be effectively resolved via schema conceptualization, structuralization and aggregation.

We conducted experiments on ERE-EN~\citep{lin-etal-2020-joint} and additional datasets in multiple domains including finance (ChFinAnn~\citep{zheng-etal-2019-doc2edag}), pandemic (Cov-19~\citep{shen-etal-2021-corpus}), and daily news (New York Time and People's Daily).
Empirical results show that ESHer surpasses the traditional methods in discovering high-quality and high-coverage event schemas.
And the induced event schemas are close to human-annotated ones and can be quickly extended to varying domains and emerging event types.

In general, this paper's main contributions are:
\begin{itemize}
\item We propose a new event schema induction paradigm -- knowledge harvesting from large-scale PLMs, which can effectively resolve the open, diversity, and sparsity challenges.
\item We design ESHer, which can automatically induce event schemas via in-context generation-based text conceptualization, confidence-aware schema structuralization, and graph-based schema aggregation.
\item Experiments show ESHer can induce high-quality and high-coverage event schemas on varying domains.
And we believe the induced event schemas are valuable resources which can benefit many downstream NLP tasks.
\end{itemize}

\begin{figure*}[!t]
\centering
\includegraphics[width=\textwidth]{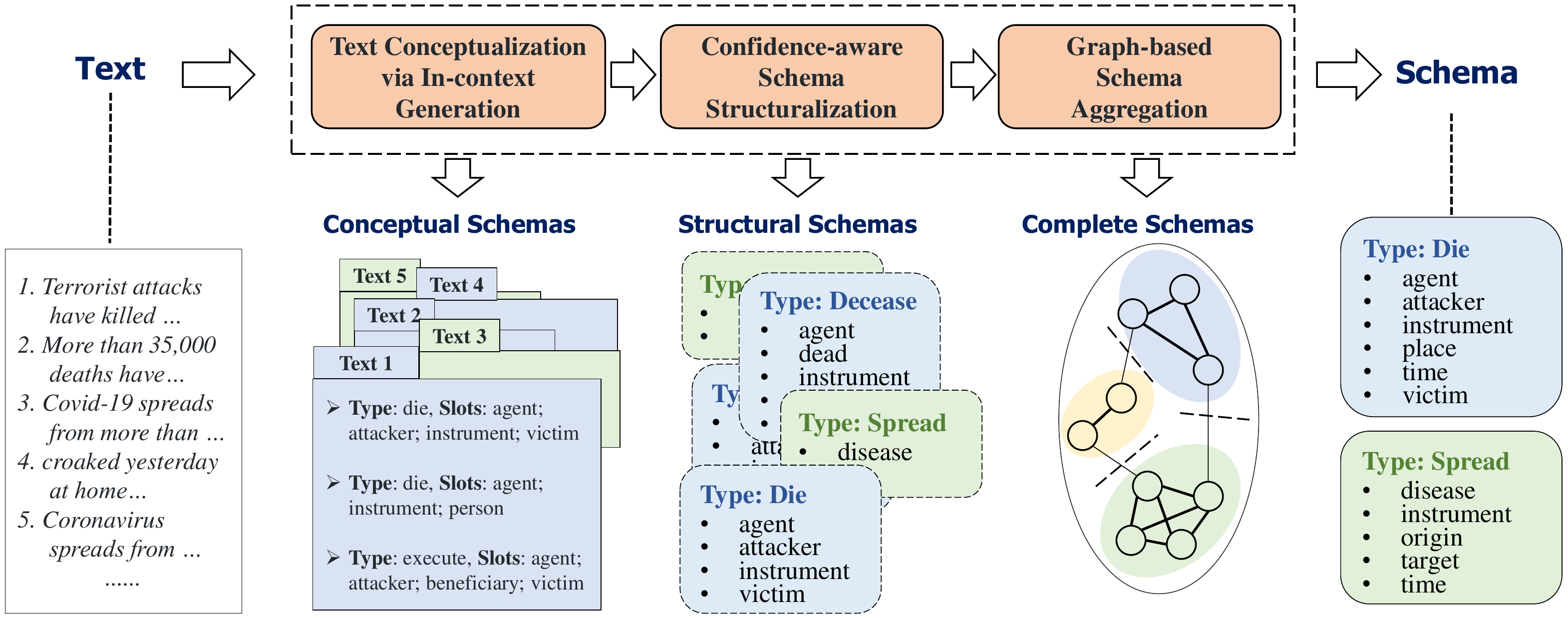}
\caption{
An overview of ESHer, which automatically discovers and normalizes event types and their semantic roles via (1) \emph{text conceptualization via in-context generation}, (2) \emph{confidence-aware schema structuralization} and (3) \emph{graph-based schema aggregation}.
}
\label{fig:esher}
\end{figure*}

\section{Event Schema Harvester} 
\label{sec:event_schema_harvester}
This section describes how to discover, conceptualize, and structuralize event schemas from large pre-trained language models (PLMs) so that the open, diversity and sparsity challenges can be effectively resolved by automatically harvesting open-domain and high-coverage event schemas from large PLMs and leveraging the strong text generation and in-context learning abilities of large PLMs.

Formally, given an unlabeled corpus $\mathcal{C}$ and a PLM, our event schema induction method discovers event clusters $\mathcal{Y}=\{y_1, y_2, ..., y_N\}$, where $N$ is the number of discovered event types.
For each event cluster $y$, we automatically conceptualize it to a name $t$ as well as its corresponding semantic roles $\{s_1^t, s_2^t, ...\}$, where $t \in \mathcal{T}$, $s \in \mathcal{S}$ and $\mathcal{T}$/$\mathcal{S}$ are open domain event type/slot names.

To this end, we design \textbf{E}vent \textbf{S}chema \textbf{H}arvest\textbf{er} (\textbf{ESHer}), and its framework is shown in Figure~\ref{fig:esher}. 
ESHer contains three components:
1) \emph{text conceptualization via in-context generation}, which transforms diverse event expressions into conceptualized event schemas based on in-context demonstrations;
2) \emph{confidence-aware schema structuralization}, which structuralizes event schemas by selecting and associating event types with their salient, reliable and consistent slots;
3) \emph{graph-based schema aggregation}, which aggregates dispersed sparse event knowledge across individual event schemas via graph-based clustering. 
In follows we describe these components in detail.

\subsection{Text Conceptualization via In-context Generation}
Events are usually expressed in diverse natural language utterances, which poses a critical challenge for schema induction.
For example, ``\emph{Terrorist attacks have killed over 35,000 people}'' and ``\emph{More than 35,000 deaths have been caused by terrorist attacks}'' convey the same event, but with quite different words and syntactic structures.
To address this challenge, we conceptualize diverse utterances into schema candidates, which can distil event schema knowledge and uniformly represent them. 
For example, our method will distil and represent the event types and the semantic roles in the above two examples as the same schema ``\emph{\textbf{Type}: die, \textbf{Slots}: agent; attacker; instrument; victim}'' (as shown in Figure~\ref{fig:esher}).

To this end, this section proposes an unsupervised text-to-schema framework by leveraging the strong in-context generation ability of PLMs. 
Specifically, we model text conceptualization as an in-context generation process:
\begin{equation}\nonumber
\begin{aligned}
[Demonstrations; Text] \rightarrow Schema
\end{aligned}
\end{equation}
where: ``\emph{Demonstrations}'' is a list of examples used to instruct PLMs how to conceptualize text to schema, and each demonstration is a <text, schema> pair represented as ``\emph{text} $\rightarrow$ \emph{schema}'', here ``\emph{Text}'' is the event utterance we want to conceptualize, ``\emph{Schema}'' is the conceptualized schema represented as ``\emph{\textbf{Type}: t, \textbf{Slots}: $s^t_1$; $s^t_2$ ...}'', and ``$\rightarrow$'' is a special token that separates the text and the event schema.

We can see that, our method is unsupervised so it can effectively resolve open and emerging events in real-world applications, and it is in-context instructed so it can be generalized to different domains/languages by instructing PLMs with appropriate in-context demonstrations.

There are many ways to select appropriate in-context demonstrations.
This paper directly samples them from existing human-annotated event datasets (e.g., ACE~\citep{doddington-etal-2004-automatic}, DuEE~\citep{li-etal-2020-duee}, etc) and we found text conceptualization benefits from high-quality and diverse demonstrations.
We believe this is because diverse demonstrations can help PLMs to better generalize to different domains, event types and event semantic roles. 
Furthermore, to recall more event knowledge from an instance, we generate $n$ schema candidates ${c_1, c_2, ... c_n}$ for each text, where $n$ is a hyperparameter.

\subsection{Confidence-aware Schema Structuralization}
The text-to-schema component distils and conceptualizes event knowledge in diverse expressions. 
This section describes how to structuralize these conceptualized event schemas by selecting and associating the salient, reliable, and consistent slots for each event type. 
For instance, we can structuralize a ``\emph{die}'' event frame by evaluating the association between event type ``\emph{die}'' and slots ``\emph{agent; attacker; instrument; victim}'' (as shown in Figure~\ref{fig:esher}). 

Formally, as shown in Algorithm~\ref{alg:aggregation}, we use $\mathcal{O}$ to denote the results of text conceptualization, in which $j$-th instance is $(text^j, \{c^j_1, c^j_2, ... c^j_n\})$ and $\{c^j_1, c^j_2, ... c^j_n\}$ are $n$ generated schema candidates, and we use $SlotSet^j$ to denote the union of all generated slots of instance j by summarizing slots in $\{c^j_1, c^j_2, ... c^j_n\}$ (\textbf{Line 2-6}).
To select high-quality slots for event types, we design a set of metrics to estimate the quality of slots and type-slot associations, including \textbf{Salience}, \textbf{Reliability} and \textbf{Consistency} (\textbf{Line 8-11}). 
We describe them as follows:

\paragraph{Salience} - a salient slot of an event type \textit{t} should appear frequently in the generated schemas of \textit{t}, but less frequent in other events. 
For example, in Figure~\ref{fig:esher}, the slots ``\emph{attacker}'' and ``\emph{victim}'' are more salient than ``\emph{person}'' for ``\emph{die}'' event. 
Following the TF-IDF idea, the salience of a slot \textit{s} in $j$-th instance is computed as:
\begin{equation}\label{equ:salience}
\begin{aligned}
Salience(s)^j &= (1+log(freq(s)^j)^2) \\
&* log(\frac{|\mathcal{O}|}{\sum^{|\mathcal{O}|}_{k} freq(s)^k})
\end{aligned}
\end{equation}
where $freq(s)^j$ is the frequency of the slot $s$ in $SlotSet^j$, $|\mathcal{O}|$ is the total number of instances in the outputs $\mathcal{O}$.

\paragraph{Reliability} - a slot is reliable if it co-occurs frequently with other slots in multiple candidates of one instance. 
For example, in Figure~\ref{fig:esher}, the slot ``\emph{agent}'' is considered reliable to ``\emph{die}'' event because it co-occurs with all other slots. 
We use PageRank algorithm~\citep{page-etal-1999-pageRank} to compute the slot reliability as follows:
\begin{equation}\label{equ:reliability}
\begin{aligned}
Reliability(s)^j &= \beta \sum_{k}^{|SlotSet^j|} \frac{Reliability(s^k)}{d(s^k)} \\
&+ (1-\beta)\frac{1}{|SlotSet^j|}
\end{aligned}
\end{equation}
where $\beta$ is a hyper-parameter, $|SlotSet^j|$ is the number of slots in $SlotSet^j$ and $d(s^k)=\sum_{k (s \leftrightarrow s^k)}^{|SlotSet^j|} Reliability(s^k)$, $s \leftrightarrow s^k$ means that slots $s$ and $s^k$ co-occur in the same candidate. 
We initialize the reliability score for all slots as $\frac{1}{|SlotSet^j|}$ and run PageRank $T$ iterations or the change is less than $\epsilon$.

{
\renewcommand\baselinestretch{1.15}
\begin{algorithm}[t]
\renewcommand{\algorithmicrequire}{\textbf{Input:}}\footnotesize
\renewcommand\algorithmicensure {\textbf{Return:} }\footnotesize
\caption{: Confidence-aware Schema Structuralization.}
\label{alg:aggregation}
\begin{algorithmic}[1]
\REQUIRE
$\mathcal{O}$: the results of text conceptualization, \\
where $j$-th instance of $\mathcal{O}$ is $(text^j, \{c^j_1, c^j_2, ... c^j_n\})$; \\
\STATE \textbf{for} $j$-th instance $\in$ $\mathcal{O}$ \textbf{do}
\STATE \ \ \ $SlotSet^j$ $\leftarrow$ $\emptyset$
\STATE \ \ \ \textbf{for} $c^j_i$ $\in$ \{$c^j_1, c^j_2, ... c^j_n$\} \textbf{do}
\STATE \ \ \ \ \ \ $c^j_i = \text{``}\emph{\textbf{Type}}: \hat{t}^{i,j} \quad \emph{\textbf{Slots}}: s^{\hat{t}^{i,j}}_1; s^{\hat{t}^{i,j}}_2; ... \text{''}$
\STATE \ \ \ \ \ \ $SlotSet^j$ $\leftarrow$ $SlotSet^j$ $\cup$ \{$s^{\hat{t}^{i,j}}_1; s^{\hat{t}^{i,j}}_2; ...$\}
\STATE \ \ \ \textbf{end for}
\STATE \ \ \ \textbf{for} $s$ $\in$ $SlotSet^j$ \textbf{do}
\STATE \ \ \ \ \ \ $Salience(s)^j$ $\leftarrow$ Equation~(\ref{equ:salience})
\STATE \ \ \ \ \ \ $Reliability(s)^j$ $\leftarrow$ Equation~(\ref{equ:reliability})
\STATE \ \ \ \ \ \ $Consistency(s)^j$ $\leftarrow$ Equation~(\ref{equ:consistency})
\STATE \ \ \ \ \ \ $Score(s)^j$ $\leftarrow$ Equation~(\ref{equ:intra})
\STATE \ \ \ \ \ \ \textbf{if} $Score(s)^j$ < threshold \textbf{do}
\STATE \ \ \ \ \ \ \ \ \ $SlotSet^j$.del($s$) 
\STATE \ \ \ \textbf{end for}
\STATE \ \ \  Select top-1 consistent event type $\hat{t}^{j}$ 
\STATE \ \ \ $j$-th instance $\leftarrow$ $(text^j, \hat{t}^{j}, SlotSet^j)$
\STATE \textbf{end for}
\ENSURE $\mathcal{O}$;
\end{algorithmic}
\end{algorithm}
\par}

\paragraph{Consistency} -- because PLMs may generate unfaithful schemas which are unfaithful to input event expressions, we also estimate the consistencies of event types and slots.
Concretely, we evaluate the consistency between the generated event schemas and event expressions using semantic similarities based on WordNet~\citep{miller-1992-wordnet}, HowNet~\citep{dong-etal-2010-hownet} and BERT~\citep{devlin-etal-2019-bert}.
And the consistency score of a slot in $j$-th instance is:
\begin{equation}\label{equ:consistency}
\begin{aligned}
Consistency(s)^j =& Sim(\hat{t}^{j}, text^j|s, \hat{t}^{j} \in c)
\end{aligned}
\end{equation}
where $Sim(\cdot)$ is a semantic similarity function, $s, \hat{t}^{j} \in o$ denotes that slot $s$ is the corresponding semantic role of the predicted event type $\hat{t}^{j}$ in the same schema candidate $c$.

The final confidence of a slot is computed by combining the salience, reliability, and consistency scores:
\begin{equation}\label{equ:intra}
\begin{aligned}
Score(s)^j &= (\lambda_1 * Salience(s)^j \\
&+ \lambda_2 * Reliability(s)^j) \\
&* Consistency(s)^j
\end{aligned}
\end{equation}
where $\lambda_1$ and $\lambda_2$ are two hyperparameters.

Finally, we only retain the top-1 consistent event type for each instance and filter all slots in that instance if their confidence scores are below a certain threshold (\textbf{Line 12-16}). In this way, we obtained structuralized event schemas such as ``\emph{\textbf{Type}: die, \textbf{Slots}: agent; attacker; instrument; victim}'' (as shown in Figure~\ref{fig:esher}).

\subsection{Graph-based Schema Aggregation}
As described above, event knowledge is sparse in event expressions due to the economical principle of language~\citep{de-2011-course}. 
This section describes how to address the sparsity issue by aggregating dispersed semantic roles across different schemas. 
For example, we can obtain a more complete schema for the ``\emph{die}'' event by combining ``\emph{\textbf{Type}: die, \textbf{Slots}: agent; attacker; instrument; victim}'' with ``\emph{\textbf{Type}: decease, \textbf{Slots}: agent; dead, instrument; place; time}''.

To this end, this section proposes a graph-based clustering method which first groups individual event schemas into clusters, and then aggregates event types and slots in the same cluster.  
The main idea here is that event schemas are of the same event type if their original expressions describe the same kind of occurrence (text similarity), their predicted types are synonyms (type similarity) and they share many common semantic slots (slot set similarity). 
For example, in Figure~\ref{fig:esher}, ``\emph{die}'' and ``\emph{decease}'' are synonyms and ``\emph{agent}'' and ``\emph{instrument}'' are common semantic roles, therefore they are highly likely the same event type.

Based on the above idea, given instances $\mathcal{O}$ after confidence-aware schema structuralization, in which the $j$-th instance is represented as $(text^j, \hat{t}^{j}, SlotSet^j)$, we construct a graph to model the similarities between different individual event schemas. 
In the graph, each node is an event schema and the similarity between two schema nodes is computed by considering the similarity between their event expressions, the similarity between their event types, and the similarity between their slot sets:
\begin{equation}\label{equ:graph}
\begin{aligned}
Graph[i][j] &= Graph[j][i] \\
&= \lambda_3 * Sim(text^i, text^j) \\
&+ \lambda_4 * Sim(\hat{t}^i, \hat{t}^j) \\
&+ \lambda_5 * Sim(SlotSet^i, SlotSet^j)
\end{aligned}
\end{equation}
where $\lambda_3$, $\lambda_4$, and $\lambda_5$ are hyper-parameters, and $Sim(\cdot)$ is the semantic similarity function defined in Equation~\ref{equ:consistency}.

Given the schema graph, we employ the Louvain algorithm~\citep{blondel-etal-2008-fast} to segment and group schemas into clusters:
\begin{equation}\label{equ:louvain}
\begin{aligned}
\hat{Y}=\{\hat{y}^1, \hat{y}^2, ..., \hat{y}^{|\mathcal{O}|}\}= Louvain(Graph)
\end{aligned}
\end{equation}
where $\hat{y}^j \in \mathcal{Y}=\{y_1, y_2, ..., y_N\}$ indicates that the $j$-th schema is assigned to the $\hat{y}^j$-th event cluster and each cluster representing a distinct event type.

Finally, we aggregate all individual event schemas in the same cluster to obtain a complete schema.
Given a cluster $y$ which can be represented as as a tuple (\emph{\textbf{Types}, \textbf{Slots}}), with $\emph{\textbf{Types}} = \{\hat{t}^1, \hat{t}^2, ...\}$ and $\emph{\textbf{Slots}} = \{s^t_1, s^t_2, ...\} = SlotSet^1 \cup SlotSet^2 \cup ...$ are the predicted event types/slots by summarizing event types/$SlotSets$ from all individual schemas.
An example of such a cluster in Figure~\ref{fig:esher} is ``\emph{(\textbf{Types}: \{die, decease\}; \textbf{Slots}: \{agent; attacker; dead; instrument; place; time; victim\})}''.

The final event type name of this cluster is normalized by selecting the most salient prediction from $\{\hat{t}^1, \hat{t}^2, ...\}$, e.g., ``\emph{die}''.
For event slots, there may be synonymous slot names in \emph{\textbf{Slots}} stand for the same semantic role, e.g., \{\emph{dead, victim}\} is the synonymous set in the above example.
Thus, we utilize the Louvain Algorithm~\citep{blondel-etal-2008-fast} again to identify synonymous event slots and then select the most salient slot to represent its synonyms, e.g., ``\emph{victim}'' is chosen as the representative slot name of the synonymous set \{\emph{dead, victim}\}\footnote{
In this step, we build a graph to model the connectivities between each slot which is similar to Equation~\ref{equ:graph}.
In the graph, each node is a slot and the similarity between two slot nodes is defined as $Sim(s^i, s^j)$.
}.
The final slot names of this cluster are normalized by these selected slots, e.g., the aggregated complete event schema of the above example is ``\emph{\textbf{Type}: die, \textbf{Slots}: agent; attacker; instrument; place; time; victim}'' (as shown in Figure~\ref{fig:esher}).

\paragraph{Summary} 
By conceptualizing diverse event expressions, structuralizing schemas by selecting and associating event types and their slots, and aggregating dispersed event knowledge across different schemas, our knowledge harvesting method can effectively address the open, diversity, and sparsity challenges, and induce conceptual, structural, and formal event schemas from PLMs.

\begin{figure*}[!t]
\centering
\includegraphics[width=\textwidth]{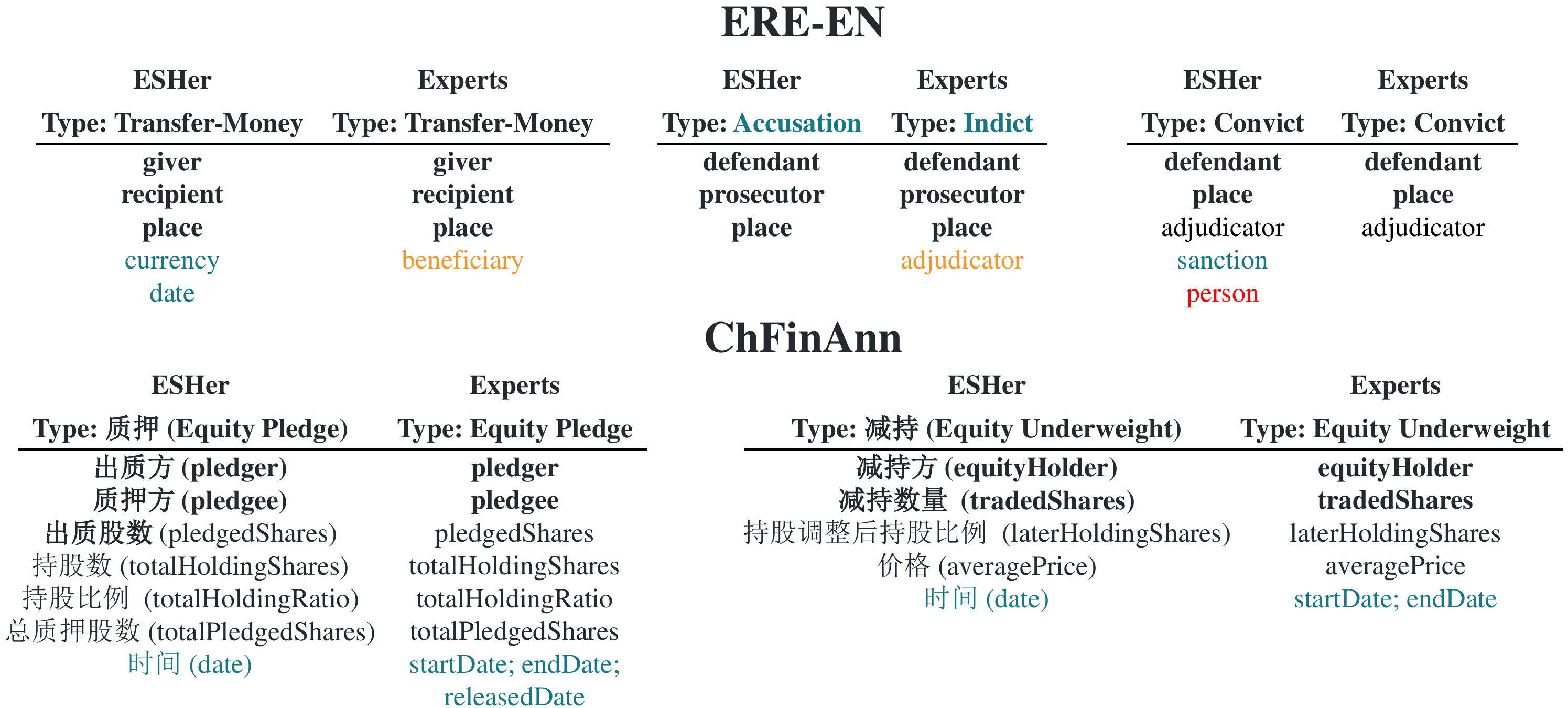}
\caption{
Schemas induced by ESHer and annotated by experts, in which \textbf{bold black} denotes the directly matched event types/slots; black denotes recalled ground truths; \textcolor{teal}{teal} denotes the unmatched but reasonable ones; \textcolor{orange}{orange} denotes the missing references; \textcolor{red}{red} denotes the wrong predictions.
}
\label{fig:discovery}
\end{figure*}

\begin{table*}[!t]
\centering
\resizebox{\textwidth}{!}{
\begin{tabular}{l|cccc|ccccc}
\hline
& \multicolumn{9}{c}{\textbf{ERE-EN}} \\
& \multicolumn{4}{c}{\textbf{\# of Event Types}} & \multicolumn{5}{c}{\textbf{\# of Event Slots}} \\
\textbf{Model} & Human & Discover & Overlap & Acceptable & Human & Discover & Overlap & Acceptable & Recall \\
\hline
ESHer & 38 & 71 & 21.05\% & 85.92\% & 115 & 198 & 11.30\% & 44.95\% & 35.21\% \\
ESHer (upper bound) & 38 & 100 & 21.05\% & 93.00\% & 115 & 371 & 19.13\% & 59.30\% & 49.00\% \\
\hline
\hline
& \multicolumn{9}{c}{\textbf{ChFinAnn}} \\
& \multicolumn{4}{c}{\textbf{\# of Event Types}} & \multicolumn{5}{c}{\textbf{\# of Event Slots}} \\
\textbf{Model} & Human & Discover & Overlap & Acceptable & Human & Discover & Overlap & Acceptable & Recall \\
ESHer & 5 & 44 & 100.00\% & 72.73\% & 35 & 231 & 37.14\% & 59.31\% & 15.91\% \\
ESHer (upper bound) & 5 & 100 & 100.00\% & 96.00\% & 35 & 458 & 71.43\% & 85.81\% & 22.00\% \\
\hline
\end{tabular}
}
\caption{
Schema Coverage Comparison on ERE-EN and ChFinAnn.
}
\label{tab:quantitative}
\end{table*}

\section{Experiments} 
\label{sec:experiments}

\subsection{Experimental Settings}

\textbf{Datasets.} 
We use ERE-EN~\citep{song-etal-2015-light} as our primary dataset because its event schemas are manually annotated. 
Furthermore, to assess event schema induction performance on different domains and languages, we further conduct experiments on various datasets including finance (ChFinAnn~\citep{zheng-etal-2019-doc2edag}), pandemic (Cov-19 and Pandemic~\citep{shen-etal-2021-corpus}) and daily news (New York Time\footnote{https://catalog.ldc.upenn.edu/LDC2008T19} and People's Daily 1946-2001\footnote{http://en.people.cn}).

\textbf{Implementation.} 
We use BLOOM~\citep{scao-etal-2022-language} in our experiments, which is a GPT-3~\citep{brown-etal-2020-language} like large-scale PLMs but is open-accessed.
For text conceptualization, we sample in-context demonstrations from ACE~\citep{doddington-etal-2004-automatic} and DuEE~\citep{li-etal-2020-duee} for both English and Chinese datasets, respectively.
The running environments and all hyper-parameters are in Appendix~\ref{sec:running_environments_and_hyper_parameters}.

\begin{figure*}[!t]
\centering
\includegraphics[width=\textwidth]{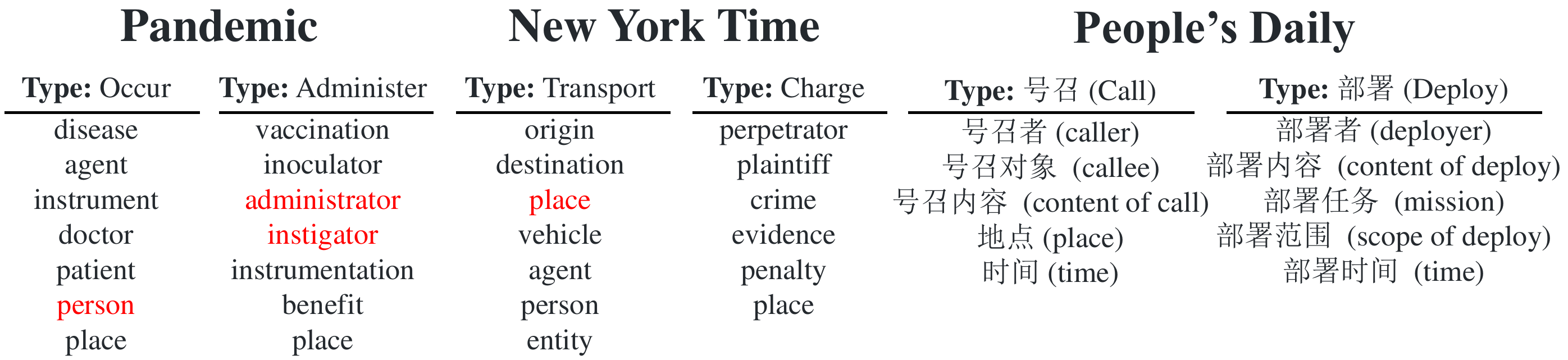}
\caption{
Event schemas induced on varying domains, in which \textcolor{black}{black} denotes the reasonable event types/slots; \textcolor{red}{red} denotes the rejected predictions.
More results can be found in Appendix~\ref{sec:more_esher_outputs}.
}
\label{fig:varying}
\end{figure*}

\subsection{Results of Event Schema Induction}
This section assesses the event schemas induced by our method. 
Following previous studies ~\citep{huang-etal-2016-liberal,shen-etal-2021-corpus}, we evaluate event schemas via the event type/slot matching task. 
Both qualitative and quantitative results show the effectiveness of the proposed ESHer. 
 
For qualitative results, Figure~\ref{fig:discovery} shows several schemas induced by ESHer from the ERE-EN and ChFinAnn datasets, and the comparison with the results of human experts is also shown. 
From Figure~\ref{fig:discovery}, we can see that ESHer can induce high-quality event schemas:
1) most of the induced event types directly match the ones annotated by experts;
2) there is a large overlap between the automatically induced slots and the manually annotated ones;
3) \textcolor{teal}{some unmatched slots} are also reasonable through our manual checking. 
This also shows that it is very hard to obtain high-coverage schemas only relying on experts.
4) we found some missing golden slots have been generated in text conceptualization but dropped in the confidence-aware structuralization step, therefore we believe the performance can be further improved by introducing human-in-the-loop.
5) with appropriate in-context demonstrations, ESHer can easily extend to different languages, e.g., English for ERE-EN and Chinese for ChFinAnn.

For quantitative results, we show the performances in Table~\ref{tab:quantitative}. We can see that:
for event type discovery, ESHer recover 21.05\% out of 38 event types in ERE-EN  and almost all (85.92\%) discovered event types are acceptable.
For event slot induction, ESHer recovers 11.30\% out of 115 slots, 44.95\% of discovered slots can be directly accepted, and 35.21\% of slots can be selected from candidates. 
This shows that event schema is a challenging task due to the diversity and sparsity of event slots.
On ChFinAnn, a typical dataset in the finance domain, we can see that ESHer is more effective and not only recover all event types but also discover lots of reasonable new types (100\% Overlap and 72.73\% Acceptable).
This shows that domain-specific event schemas can be better induced, we believe this may be because domain-specific events are more salient in domain corpus.
To assess the performance of graph-based schema aggregation, we manually check 100 individual schemas and cluster them, and its performance is shown as ESHer (upper bound) which can be regarded as the upper bound of graph-based schema aggregation. 
We can see that the performance gap between ESHer and ESHer (upper bound) is not large, which verifies the effectiveness of our graph-based schema aggregation component.

\begin{table}[!t]
\centering
\resizebox{\columnwidth}{!}{
\begin{tabular}{l|ccc}  
\hline 
\textbf{Methods} & \textbf{\quad ARI \quad} & \textbf{\quad NMI \quad} & \textbf{BCubed-F1} \\
\hline 
\hline 
Kmeans & 12.51 & 37.65 & 31.01 \\
AggClus & 13.11 & 39.16 & 31.20 \\
JCSC & 17.69 & 43.40 &  37.64 \\
Triframes-CW & 5.79 & 25.73 & 33.61 \\
Triframes-Watset & 7.53 & 47.43 & 24.04 \\
ETypeClus & 10.18 & 36.17 & 28.99 \\
\hline 
\hline 
ESHer & \textbf{56.59} & \textbf{67.72}  & \textbf{62.43}  \\
\hline 
\quad - Salience & 32.84 & 57.91 & 52.51 \\
\quad - Reliability & 52.54 & 63.86 & 61.47 \\
\quad - Consistency & 37.51 & 66.12 & 50.69 \\
\quad only Salience & 38.75 & 66.34 & 50.69 \\
\quad only Reliability & 33.98 & 62.43 & 51.09 \\
\hline 
\end{tabular}
}
\caption{
Event mention clustering results on ERE-EN. 
All values are in percentage. 
We run each method 10 times and report its averaged result for each metric. 
Note that for ESHer and its variants, due to the huge computing cost, we only run them once.
}
\label{tab:cluster_results}
\end{table}

\subsection{Results on Event Mention Clustering}
We also evaluate the effectiveness of ESHer via the event mention clustering task. 
Following~\citet{shen-etal-2021-corpus}, we select 15 event types with the most mentions and cluster all candidates into several groups for ERE-EN.

\textbf{Evaluation Metrics.}
To evaluate whether clusters (Equation~\ref{equ:louvain}) align well with the original types, we choose several standard metrics:
1) \textbf{ARI}~\citep{hubert-etal-1985-comparing} measures the similarity;
2) \textbf{NMI} measures the normalized mutual information;
3) \textbf{BCubed-F1}~\citep{bagga-baldwin-1998-entity-based} measures the aggregated precision and recall. 
For all the above metrics, the higher the values, the better the model performance.
The math formulas of these metrics are in Appendix ~\ref{sec:evaluation_metrics_for_event_mention_clustering}.

\textbf{Baselines.}
We compare ESHer with the following feature-based methods: \emph{Kmeans}, \emph{AggClus}, \emph{JCSC}~\citep{huang-etal-2016-liberal}, \emph{Triframes-CW}~\citep{ustalov-etal-2018-unsupervised-semantic}, \emph{Triframes-Watset}~\citep{ustalov-etal-2018-unsupervised-semantic} and \emph{ETypeClus}~\citep{shen-etal-2021-corpus}.
We set all hyper-parameters of these clusters using the default settings of~\citet{shen-etal-2021-corpus}.

\textbf{Experimental Results.}
Table~\ref{tab:cluster_results} shows the overall results.
For our approach, we use the full ESHer and its five ablated settings:
ESHer-Consistency, ESHer-Salience, ESHer-Reliability, ESHer only Reliability and ESHer only Salience, where Salience, Reliability and Consistency denote different estimations described in confidence-aware schema structuralization.
We can see that:

\textbf{1. ESHer outperforms all baselines on ERE-EN on all metrics.}
ESHer achieves state-of-the-art performance with 56.59 ARI, 67.72 NMI and 62.43 BCubed-F1.
We believe this is because ESHer fully leverages the in-context learning ability of PLMs, therefore the diverse and sparse challenges can be effectively resolved.

\textbf{2. The proposed salience, reliability and consistency estimations are all useful and complementary to each other.}
Compared with the full ESHer model, all five variants show declined performance to different degrees.
ESHer outperforms ESHer-Salience 23.75 ARI, 9.81 NMI and 9.92 BCubed-F1, and this result verifies the effectiveness of the salience score for identifying good slots.
ESHer outperforms ESHer-Consistency 19.08 ARI, 1.60 NMI and 11.74 BCubed-F1, this shows that consistency estimation is also indispensable.
These results also verify that high-quality slot sets are beneficial for graph-based aggregation.

\subsection{Results on Different Domains}
This section assesses event schemas induced on different domains such as pandemic and daily news. 
Figure~\ref{fig:varying} shows the result schemas and we can see that ESHer is robust in different domains and can be generalized in different settings.
Furthermore, the results also present challenges:
1) \textbf{the granularity alignment problem}: the slots in the same schema may have different granularities, e.g., the ``\emph{\textcolor{red}{person}}'' and ``\emph{doctor, patient}'' in Schema 1 on the pandemic domain;
2) \textbf{the polysemy problem}: event type ``\emph{administer}'' in Schema 2 on pandemic domain misdirects the slot ``\emph{\textcolor{red}{administrator}}'';
3) \textbf{emotional expressions}: event schema knowledge should be objective but ``\emph{\textcolor{red}{instigator}}'' conveys negative sentiment.

\section{Related Work}
\label{sec:related_work}

\textbf{Event Schema Induction.}
Traditional event schema induction methods mostly mine event schemas from raw corpora, and two main categories of methods have been proposed: 
Bottom-up concept linking methods~\citep{huang-etal-2016-liberal,he-etal-2022-acquiring} discover event types/slots by parsing and linking event expressions to external schema resources such as FrameNet~\citep{baker-etal-1998-berkeley-framenet};
Top-down clustering methods~\citep{chambers-2013-event,cheung-etal-2013-probabilistic,nguyen-etal-2015-generative,sha-etal-2016-joint,ahn-2017-inducing,yuan-etal-2018-open,liu-etal-2019-open,shen-etal-2021-corpus} cluster event expressions according to pre-defined schema templates (e.g., the 5W1H template, or templates with the predefined number of event types/slots).
Most recently, \citet{li-etal-2022-open} train a model to predict discovered event type names.
Differently, this paper induces formal event schemas in unsupervised behaviour.

There were also some other studies such as script learning~\citep{chambers-jurafsky-2008-unsupervised,chambers-jurafsky-2009-unsupervised,pichotta-mooney-2014-statistical} and event graph schema induction~\citep{li-etal-2021-future,zhao-etal-2022-connecting,du-etal-2022-resin}, which focus on mining event relations and narrative schemas. 
This paper doesn't address these issues and leaves them as future works.

\textbf{Harvesting Knowledge from Large-scale Language Models.}
Large-scale pre-trained language models (PLMs) such as GPT-3~\citep{brown-etal-2020-language} and BLOOM~\citep{scao-etal-2022-language} have been verified containing massive knowledge such as linguistic knowledge~\citep{liu-etal-2019-linguistic,warstadt-etal-2020-blimp-benchmark}, factual knowledge~\citep{petroni-etal-2019-language}, commonsense knowledge~\citep{zhou-etal-2020-evaluating} and reasoning knowledge~\citep{talmor-etal-2020-olmpics}.
Furthermore, PLMs also have shown many emergent abilities~\citep{wei-etal-2022-chain} such as in-context learning, chain-of-thought reasoning, etc. 
In recent years, researchers start to learn how to harvest resources from PLMs, such as knowledge graphs~\citep{west-etal-2022-symbolic} and explanation datasets~\citep{li-etal-2022-explanations}. 
In this paper, we study how to harvest open-domain, high-quality and high-coverage event schemas from PLMs by leveraging the abilities of PLMs.

\section{Conclusions}
\label{sec:conclusions}
In this paper, we propose a new paradigm for event schema induction -- knowledge harvesting from pre-trained language models (PLMs), which can effectively resolve the open, diversity and sparsity challenges by discovering, conceptualizing and structuralizing event schemas from PLMs. 
And an \textbf{E}vent \textbf{S}chema \textbf{H}arvest\textbf{er} (\textbf{ESHer}) is designed to automatically induce high-quality event schemas.
Empirical results show that ESHer can induce high-quality and high-coverage event schemas on different domains. 
Event schemas are valuable resources, we want to harvest and release an open, large-scale event schema repository to research communities. 

\section*{Limitations}
In this paper, we harvest event schemas from large pre-trained language models.
However, the huge cost of computing resources cannot be ignored.
Therefore, in future work, we want to study how to distil open-domain, in-context learning abilities from massive language models to smaller ones.

Second, some outputs of large language models may be biased, e.g., containing mentions of discriminatory or offensive language.
Thus, in practice, we can introduce human-in-the-loop to avoid these concerns.

\section*{Ethics Statement}
In consideration of ethical concerns, we provide the following detailed description:
\begin{itemize}
\item All of the datasets used in this paper come from Linguistic Data Consortium (LDC)\footnote{https://www.ldc.upenn.edu/}, previous studies~\citep{shen-etal-2021-corpus} and publicly available sources.
LDC datasets are accessible online for academic use with corresponding licenses and other datasets are freely accessible online without copyright constraints.
\item The large-scale pre-trained language model BLOOM~\citep{scao-etal-2022-language} is open-accessed in Huggingface Model Collections~\citep{wolf-etal-2020-transformers}.
\end{itemize}

\bibliography{acl2023}

\begin{thebibliography}{54}
\expandafter\ifx\csname natexlab\endcsname\relax\def\natexlab#1{#1}\fi

\bibitem[{Ahn(2017)}]{ahn-2017-inducing}
Natalie Ahn. 2017.
\newblock \href {https://doi.org/10.18653/v1/W17-2710} {Inducing event types
  and roles in reverse: Using function to discover theme}.
\newblock In \emph{Proceedings of the Events and Stories in the News Workshop},
  pages 66--76, Vancouver, Canada. Association for Computational Linguistics.

\bibitem[{Attardi(2015)}]{giusepppe-2015-wikiextractor}
Giusepppe Attardi. 2015.
\newblock Wikiextractor.
\newblock https://github.com/attardi/wikiextractor.

\bibitem[{Bagga and Baldwin(1998)}]{bagga-baldwin-1998-entity-based}
Amit Bagga and Breck Baldwin. 1998.
\newblock \href {https://doi.org/10.3115/980845.980859} {Entity-based
  cross-document coreferencing using the vector space model}.
\newblock In \emph{36th Annual Meeting of the Association for Computational
  Linguistics and 17th International Conference on Computational Linguistics,
  Volume 1}, pages 79--85, Montreal, Quebec, Canada. Association for
  Computational Linguistics.

\bibitem[{Baker et~al.(1998)Baker, Fillmore, and
  Lowe}]{baker-etal-1998-berkeley-framenet}
Collin~F. Baker, Charles~J. Fillmore, and John~B. Lowe. 1998.
\newblock \href {https://doi.org/10.3115/980845.980860} {The {B}erkeley
  {F}rame{N}et project}.
\newblock In \emph{36th Annual Meeting of the Association for Computational
  Linguistics and 17th International Conference on Computational Linguistics,
  Volume 1}, pages 86--90, Montreal, Quebec, Canada. Association for
  Computational Linguistics.

\bibitem[{Blondel et~al.(2008)Blondel, Guillaume, Lambiotte, and
  Lefebvre}]{blondel-etal-2008-fast}
Vincent~D Blondel, Jean-Loup Guillaume, Renaud Lambiotte, and Etienne Lefebvre.
  2008.
\newblock Fast unfolding of communities in large networks.
\newblock \emph{Journal of statistical mechanics: theory and experiment}.

\bibitem[{Brown et~al.(2020)Brown, Mann, Ryder, Subbiah, Kaplan, Dhariwal,
  Neelakantan, Shyam, Sastry, Askell et~al.}]{brown-etal-2020-language}
Tom Brown, Benjamin Mann, Nick Ryder, Melanie Subbiah, Jared~D Kaplan, Prafulla
  Dhariwal, Arvind Neelakantan, Pranav Shyam, Girish Sastry, Amanda Askell,
  et~al. 2020.
\newblock Language models are few-shot learners.
\newblock \emph{Advances in neural information processing systems}.

\bibitem[{Chambers(2013)}]{chambers-2013-event}
Nathanael Chambers. 2013.
\newblock \href {https://aclanthology.org/D13-1185} {Event schema induction
  with a probabilistic entity-driven model}.
\newblock In \emph{Proceedings of the 2013 Conference on Empirical Methods in
  Natural Language Processing}, pages 1797--1807, Seattle, Washington, USA.
  Association for Computational Linguistics.

\bibitem[{Chambers and Jurafsky(2008)}]{chambers-jurafsky-2008-unsupervised}
Nathanael Chambers and Dan Jurafsky. 2008.
\newblock \href {https://aclanthology.org/P08-1090} {Unsupervised learning of
  narrative event chains}.
\newblock In \emph{Proceedings of ACL-08: HLT}, pages 789--797, Columbus, Ohio.
  Association for Computational Linguistics.

\bibitem[{Chambers and Jurafsky(2009)}]{chambers-jurafsky-2009-unsupervised}
Nathanael Chambers and Dan Jurafsky. 2009.
\newblock \href {https://aclanthology.org/P09-1068} {Unsupervised learning of
  narrative schemas and their participants}.
\newblock In \emph{Proceedings of the Joint Conference of the 47th Annual
  Meeting of the {ACL} and the 4th International Joint Conference on Natural
  Language Processing of the {AFNLP}}, pages 602--610, Suntec, Singapore.
  Association for Computational Linguistics.

\bibitem[{Chambers and Jurafsky(2011)}]{chambers-jurafsky-2011-template}
Nathanael Chambers and Dan Jurafsky. 2011.
\newblock \href {https://aclanthology.org/P11-1098} {Template-based information
  extraction without the templates}.
\newblock In \emph{Proceedings of the 49th Annual Meeting of the Association
  for Computational Linguistics: Human Language Technologies}, pages 976--986,
  Portland, Oregon, USA. Association for Computational Linguistics.

\bibitem[{Cheung et~al.(2013)Cheung, Poon, and
  Vanderwende}]{cheung-etal-2013-probabilistic}
Jackie Chi~Kit Cheung, Hoifung Poon, and Lucy Vanderwende. 2013.
\newblock \href {https://aclanthology.org/N13-1104} {Probabilistic frame
  induction}.
\newblock In \emph{Proceedings of the 2013 Conference of the North {A}merican
  Chapter of the Association for Computational Linguistics: Human Language
  Technologies}, pages 837--846, Atlanta, Georgia. Association for
  Computational Linguistics.

\bibitem[{Chinchor et~al.(1993)Chinchor, Hirschman, and
  Lewis}]{chinchor-etal-1993-evaluating}
Nancy Chinchor, Lynette Hirschman, and David~D. Lewis. 1993.
\newblock \href {https://aclanthology.org/J93-3001} {Evaluating message
  understanding systems: An analysis of the third {M}essage {U}nderstanding
  {C}onference ({MUC}-3)}.
\newblock \emph{Computational Linguistics}, 19(3):409--450.

\bibitem[{De~Saussure(2011)}]{de-2011-course}
Ferdinand De~Saussure. 2011.
\newblock \emph{Course in General Linguistics}.
\newblock Columbia University Press.

\bibitem[{Devlin et~al.(2019)Devlin, Chang, Lee, and
  Toutanova}]{devlin-etal-2019-bert}
Jacob Devlin, Ming-Wei Chang, Kenton Lee, and Kristina Toutanova. 2019.
\newblock \href {https://doi.org/10.18653/v1/N19-1423} {{BERT}: Pre-training of
  deep bidirectional transformers for language understanding}.
\newblock In \emph{Proceedings of the 2019 Conference of the North {A}merican
  Chapter of the Association for Computational Linguistics: Human Language
  Technologies, Volume 1 (Long and Short Papers)}, pages 4171--4186,
  Minneapolis, Minnesota. Association for Computational Linguistics.

\bibitem[{Doddington et~al.(2004)Doddington, Mitchell, Przybocki, Ramshaw,
  Strassel, and Weischedel}]{doddington-etal-2004-automatic}
George Doddington, Alexis Mitchell, Mark Przybocki, Lance Ramshaw, Stephanie
  Strassel, and Ralph Weischedel. 2004.
\newblock \href {http://www.lrec-conf.org/proceedings/lrec2004/pdf/5.pdf} {The
  automatic content extraction ({ACE}) program {--} tasks, data, and
  evaluation}.
\newblock In \emph{Proceedings of the Fourth International Conference on
  Language Resources and Evaluation ({LREC}{'}04)}, Lisbon, Portugal. European
  Language Resources Association (ELRA).

\bibitem[{Dong et~al.(2010)Dong, Dong, and Hao}]{dong-etal-2010-hownet}
Zhendong Dong, Qiang Dong, and Changling Hao. 2010.
\newblock \href {https://aclanthology.org/C10-3014} {{H}ow{N}et and its
  computation of meaning}.
\newblock In \emph{Coling 2010: Demonstrations}, pages 53--56, Beijing, China.
  Coling 2010 Organizing Committee.

\bibitem[{Du et~al.(2022)Du, Zhang, Li, Yu, Wang, Lai, Lin, Wang, Liu, Zhou,
  Wen, Li, Hannan, Lei, Kim, Dror, Wang, Regan, Zeng, Lyu, Yu, Edwards, Jin,
  Jiao, Kazeminejad, Wang, Callison-Burch, Bansal, Vondrick, Han, Roth, Chang,
  Palmer, and Ji}]{du-etal-2022-resin}
Xinya Du, Zixuan Zhang, Sha Li, Pengfei Yu, Hongwei Wang, Tuan Lai, Xudong Lin,
  Ziqi Wang, Iris Liu, Ben Zhou, Haoyang Wen, Manling Li, Darryl Hannan, Jie
  Lei, Hyounghun Kim, Rotem Dror, Haoyu Wang, Michael Regan, Qi~Zeng, Qing Lyu,
  Charles Yu, Carl Edwards, Xiaomeng Jin, Yizhu Jiao, Ghazaleh Kazeminejad,
  Zhenhailong Wang, Chris Callison-Burch, Mohit Bansal, Carl Vondrick, Jiawei
  Han, Dan Roth, Shih-Fu Chang, Martha Palmer, and Heng Ji. 2022.
\newblock \href {https://doi.org/10.18653/v1/2022.naacl-demo.7} {{RESIN}-11:
  Schema-guided event prediction for 11 newsworthy scenarios}.
\newblock In \emph{Proceedings of the 2022 Conference of the North American
  Chapter of the Association for Computational Linguistics: Human Language
  Technologies: System Demonstrations}, pages 54--63, Hybrid: Seattle,
  Washington + Online. Association for Computational Linguistics.

\bibitem[{He et~al.(2022)He, Fang, Wang, and Song}]{he-etal-2022-acquiring}
Mutian He, Tianqing Fang, Weiqi Wang, and Yangqiu Song. 2022.
\newblock Acquiring and modelling abstract commonsense knowledge via
  conceptualization.
\newblock \emph{arXiv preprint arXiv:2206.01532}.

\bibitem[{Huang et~al.(2016)Huang, Cassidy, Feng, Ji, Voss, Han, and
  Sil}]{huang-etal-2016-liberal}
Lifu Huang, Taylor Cassidy, Xiaocheng Feng, Heng Ji, Clare~R. Voss, Jiawei Han,
  and Avirup Sil. 2016.
\newblock \href {https://doi.org/10.18653/v1/P16-1025} {Liberal event
  extraction and event schema induction}.
\newblock In \emph{Proceedings of the 54th Annual Meeting of the Association
  for Computational Linguistics (Volume 1: Long Papers)}, pages 258--268,
  Berlin, Germany. Association for Computational Linguistics.

\bibitem[{Hubert and Arabie(1985)}]{hubert-etal-1985-comparing}
Lawrence Hubert and Phipps Arabie. 1985.
\newblock Comparing partitions.
\newblock \emph{Journal of classification}.

\bibitem[{Irwin et~al.(2011)Irwin, Komachi, and
  Matsumoto}]{irwin-etal-2011-narrative}
Joseph Irwin, Mamoru Komachi, and Yuji Matsumoto. 2011.
\newblock \href {https://aclanthology.org/W11-1913} {Narrative schema as world
  knowledge for coreference resolution}.
\newblock In \emph{Proceedings of the Fifteenth Conference on Computational
  Natural Language Learning: Shared Task}, pages 86--92, Portland, Oregon, USA.
  Association for Computational Linguistics.

\bibitem[{Jackendoff(1992)}]{jackendoff-1992-semantic}
Ray~S Jackendoff. 1992.
\newblock \emph{Semantic structures}.
\newblock MIT press.

\bibitem[{Ji and Grishman(2011)}]{ji-grishman-2011-knowledge}
Heng Ji and Ralph Grishman. 2011.
\newblock \href {https://aclanthology.org/P11-1115} {Knowledge base population:
  Successful approaches and challenges}.
\newblock In \emph{Proceedings of the 49th Annual Meeting of the Association
  for Computational Linguistics: Human Language Technologies}, pages
  1148--1158, Portland, Oregon, USA. Association for Computational Linguistics.

\bibitem[{Li et~al.(2021)Li, Li, Wang, Huang, Cho, Ji, Han, and
  Voss}]{li-etal-2021-future}
Manling Li, Sha Li, Zhenhailong Wang, Lifu Huang, Kyunghyun Cho, Heng Ji,
  Jiawei Han, and Clare Voss. 2021.
\newblock \href {https://doi.org/10.18653/v1/2021.emnlp-main.422} {The future
  is not one-dimensional: Complex event schema induction by graph modeling for
  event prediction}.
\newblock In \emph{Proceedings of the 2021 Conference on Empirical Methods in
  Natural Language Processing}, pages 5203--5215, Online and Punta Cana,
  Dominican Republic. Association for Computational Linguistics.

\bibitem[{Li et~al.(2020{\natexlab{a}})Li, Zeng, Lin, Cho, Ji, May, Chambers,
  and Voss}]{li-etal-2020-connecting}
Manling Li, Qi~Zeng, Ying Lin, Kyunghyun Cho, Heng Ji, Jonathan May, Nathanael
  Chambers, and Clare Voss. 2020{\natexlab{a}}.
\newblock \href {https://doi.org/10.18653/v1/2020.emnlp-main.50} {Connecting
  the dots: Event graph schema induction with path language modeling}.
\newblock In \emph{Proceedings of the 2020 Conference on Empirical Methods in
  Natural Language Processing (EMNLP)}, pages 684--695, Online. Association for
  Computational Linguistics.

\bibitem[{Li et~al.(2022{\natexlab{a}})Li, Ji, and Han}]{li-etal-2022-open}
Sha Li, Heng Ji, and Jiawei Han. 2022{\natexlab{a}}.
\newblock Open relation and event type discovery with type abstraction.
\newblock In \emph{Proceedings of the 2022 Conference on Empirical Methods in
  Natural Language Processing}.

\bibitem[{Li et~al.(2022{\natexlab{b}})Li, Chen, Shen, Chen, Zhang, Li, Wang,
  Qian, Peng, Mao et~al.}]{li-etal-2022-explanations}
Shiyang Li, Jianshu Chen, Yelong Shen, Zhiyu Chen, Xinlu Zhang, Zekun Li, Hong
  Wang, Jing Qian, Baolin Peng, Yi~Mao, et~al. 2022{\natexlab{b}}.
\newblock Explanations from large language models make small reasoners better.
\newblock \emph{arXiv preprint arXiv:2210.06726}.

\bibitem[{Li et~al.(2020{\natexlab{b}})Li, Li, Pan, Chen, Peng, Wang, Lyu, and
  Z~hu}]{li-etal-2020-duee}
Xinyu Li, Fayuan Li, Lu~Pan, Yuguang Chen, Weihua Peng, Quan Wang, Yajuan Lyu,
  and Yong Z~hu. 2020{\natexlab{b}}.
\newblock Duee: a large-scale dataset for chinese event extraction in
  real-world scenarios.
\newblock In \emph{CCF International Conference on Natural Language Processing
  and Chinese Computing}.

\bibitem[{Lin et~al.(2020)Lin, Ji, Huang, and Wu}]{lin-etal-2020-joint}
Ying Lin, Heng Ji, Fei Huang, and Lingfei Wu. 2020.
\newblock \href {https://doi.org/10.18653/v1/2020.acl-main.713} {A joint neural
  model for information extraction with global features}.
\newblock In \emph{Proceedings of the 58th Annual Meeting of the Association
  for Computational Linguistics}, pages 7999--8009, Online. Association for
  Computational Linguistics.

\bibitem[{Liu et~al.(2019{\natexlab{a}})Liu, Gardner, Belinkov, Peters, and
  Smith}]{liu-etal-2019-linguistic}
Nelson~F. Liu, Matt Gardner, Yonatan Belinkov, Matthew~E. Peters, and Noah~A.
  Smith. 2019{\natexlab{a}}.
\newblock \href {https://doi.org/10.18653/v1/N19-1112} {Linguistic knowledge
  and transferability of contextual representations}.
\newblock In \emph{Proceedings of the 2019 Conference of the North {A}merican
  Chapter of the Association for Computational Linguistics: Human Language
  Technologies, Volume 1 (Long and Short Papers)}, pages 1073--1094,
  Minneapolis, Minnesota. Association for Computational Linguistics.

\bibitem[{Liu et~al.(2019{\natexlab{b}})Liu, Huang, and
  Zhang}]{liu-etal-2019-open}
Xiao Liu, Heyan Huang, and Yue Zhang. 2019{\natexlab{b}}.
\newblock \href {https://doi.org/10.18653/v1/P19-1276} {Open domain event
  extraction using neural latent variable models}.
\newblock In \emph{Proceedings of the 57th Annual Meeting of the Association
  for Computational Linguistics}, pages 2860--2871, Florence, Italy.
  Association for Computational Linguistics.

\bibitem[{Lu et~al.(2021)Lu, Lin, Xu, Han, Tang, Li, Sun, Liao, and
  Chen}]{lu-etal-2021-text2event}
Yaojie Lu, Hongyu Lin, Jin Xu, Xianpei Han, Jialong Tang, Annan Li, Le~Sun,
  Meng Liao, and Shaoyi Chen. 2021.
\newblock \href {https://doi.org/10.18653/v1/2021.acl-long.217}
  {{T}ext2{E}vent: Controllable sequence-to-structure generation for end-to-end
  event extraction}.
\newblock In \emph{Proceedings of the 59th Annual Meeting of the Association
  for Computational Linguistics and the 11th International Joint Conference on
  Natural Language Processing (Volume 1: Long Papers)}, pages 2795--2806,
  Online. Association for Computational Linguistics.

\bibitem[{Lu et~al.(2022)Lu, Liu, Dai, Xiao, Lin, Han, Sun, and
  Wu}]{lu-etal-2022-unified}
Yaojie Lu, Qing Liu, Dai Dai, Xinyan Xiao, Hongyu Lin, Xianpei Han, Le~Sun, and
  Hua Wu. 2022.
\newblock \href {https://doi.org/10.18653/v1/2022.acl-long.395} {Unified
  structure generation for universal information extraction}.
\newblock In \emph{Proceedings of the 60th Annual Meeting of the Association
  for Computational Linguistics (Volume 1: Long Papers)}, pages 5755--5772,
  Dublin, Ireland. Association for Computational Linguistics.

\bibitem[{Miller(1992)}]{miller-1992-wordnet}
George~A. Miller. 1992.
\newblock \href {https://aclanthology.org/H92-1116} {{W}ord{N}et: A lexical
  database for {E}nglish}.
\newblock In \emph{Speech and Natural Language: Proceedings of a Workshop Held
  at Harriman, New York, {F}ebruary 23-26, 1992}.

\bibitem[{Nguyen et~al.(2015)Nguyen, Tannier, Ferret, and
  Besan{\c{c}}on}]{nguyen-etal-2015-generative}
Kiem-Hieu Nguyen, Xavier Tannier, Olivier Ferret, and Romaric Besan{\c{c}}on.
  2015.
\newblock \href {https://doi.org/10.3115/v1/P15-1019} {Generative event schema
  induction with entity disambiguation}.
\newblock In \emph{Proceedings of the 53rd Annual Meeting of the Association
  for Computational Linguistics and the 7th International Joint Conference on
  Natural Language Processing (Volume 1: Long Papers)}, pages 188--197,
  Beijing, China. Association for Computational Linguistics.

\bibitem[{Page et~al.(1999)Page, Brin, Motwani, and
  Winograd}]{page-etal-1999-pageRank}
Lawrence Page, Sergey Brin, Rajeev Motwani, and Terry Winograd. 1999.
\newblock The pagerank citation ranking: Bringing order to the web.
\newblock Technical report, Stanford InfoLab.

\bibitem[{Pedregosa et~al.(2011)Pedregosa, Varoquaux, Gramfort, Michel,
  Thirion, Grisel, Blondel, Prettenhofer, Weiss, Dubourg
  et~al.}]{pedregosa-etal-2011-scikit}
Fabian Pedregosa, Ga{\"e}l Varoquaux, Alexandre Gramfort, Vincent Michel,
  Bertrand Thirion, Olivier Grisel, Mathieu Blondel, Peter Prettenhofer, Ron
  Weiss, Vincent Dubourg, et~al. 2011.
\newblock Scikit-learn: Machine learning in python.
\newblock \emph{the Journal of machine Learning research}.

\bibitem[{Petroni et~al.(2019)Petroni, Rockt{\"a}schel, Riedel, Lewis, Bakhtin,
  Wu, and Miller}]{petroni-etal-2019-language}
Fabio Petroni, Tim Rockt{\"a}schel, Sebastian Riedel, Patrick Lewis, Anton
  Bakhtin, Yuxiang Wu, and Alexander Miller. 2019.
\newblock \href {https://doi.org/10.18653/v1/D19-1250} {Language models as
  knowledge bases?}
\newblock In \emph{Proceedings of the 2019 Conference on Empirical Methods in
  Natural Language Processing and the 9th International Joint Conference on
  Natural Language Processing (EMNLP-IJCNLP)}, pages 2463--2473, Hong Kong,
  China. Association for Computational Linguistics.

\bibitem[{Pichotta and Mooney(2014)}]{pichotta-mooney-2014-statistical}
Karl Pichotta and Raymond Mooney. 2014.
\newblock \href {https://doi.org/10.3115/v1/E14-1024} {Statistical script
  learning with multi-argument events}.
\newblock In \emph{Proceedings of the 14th Conference of the {E}uropean Chapter
  of the Association for Computational Linguistics}, pages 220--229,
  Gothenburg, Sweden. Association for Computational Linguistics.

\bibitem[{Scao et~al.(2022)Scao, Wang, Hesslow, Saulnier, Bekman, Bari,
  Bideman, Elsahar, Muennighoff, Phang et~al.}]{scao-etal-2022-language}
Teven~Le Scao, Thomas Wang, Daniel Hesslow, Lucile Saulnier, Stas Bekman,
  M~Saiful Bari, Stella Bideman, Hady Elsahar, Niklas Muennighoff, Jason Phang,
  et~al. 2022.
\newblock What language model to train if you have one million gpu hours?
\newblock \emph{arXiv preprint arXiv:2210.15424}.

\bibitem[{Sha et~al.(2016)Sha, Li, Chang, and Sui}]{sha-etal-2016-joint}
Lei Sha, Sujian Li, Baobao Chang, and Zhifang Sui. 2016.
\newblock \href {https://doi.org/10.18653/v1/N16-1049} {Joint learning
  templates and slots for event schema induction}.
\newblock In \emph{Proceedings of the 2016 Conference of the North {A}merican
  Chapter of the Association for Computational Linguistics: Human Language
  Technologies}, pages 428--434, San Diego, California. Association for
  Computational Linguistics.

\bibitem[{Shen et~al.(2021)Shen, Zhang, Ji, and Han}]{shen-etal-2021-corpus}
Jiaming Shen, Yunyi Zhang, Heng Ji, and Jiawei Han. 2021.
\newblock \href {https://doi.org/10.18653/v1/2021.emnlp-main.441} {Corpus-based
  open-domain event type induction}.
\newblock In \emph{Proceedings of the 2021 Conference on Empirical Methods in
  Natural Language Processing}, pages 5427--5440, Online and Punta Cana,
  Dominican Republic. Association for Computational Linguistics.

\bibitem[{Song et~al.(2015)Song, Bies, Strassel, Riese, Mott, Ellis, Wright,
  Kulick, Ryant, and Ma}]{song-etal-2015-light}
Zhiyi Song, Ann Bies, Stephanie Strassel, Tom Riese, Justin Mott, Joe Ellis,
  Jonathan Wright, Seth Kulick, Neville Ryant, and Xiaoyi Ma. 2015.
\newblock \href {https://doi.org/10.3115/v1/W15-0812} {From light to rich
  {ERE}: Annotation of entities, relations, and events}.
\newblock In \emph{Proceedings of the The 3rd Workshop on {EVENTS}: Definition,
  Detection, Coreference, and Representation}, pages 89--98, Denver, Colorado.
  Association for Computational Linguistics.

\bibitem[{Talmor et~al.(2020)Talmor, Elazar, Goldberg, and
  Berant}]{talmor-etal-2020-olmpics}
Alon Talmor, Yanai Elazar, Yoav Goldberg, and Jonathan Berant. 2020.
\newblock \href {https://doi.org/10.1162/tacl_a_00342} {o{LM}pics-on what
  language model pre-training captures}.
\newblock \emph{Transactions of the Association for Computational Linguistics},
  8:743--758.

\bibitem[{Ustalov et~al.(2018)Ustalov, Panchenko, Kutuzov, Biemann, and
  Ponzetto}]{ustalov-etal-2018-unsupervised-semantic}
Dmitry Ustalov, Alexander Panchenko, Andrey Kutuzov, Chris Biemann, and
  Simone~Paolo Ponzetto. 2018.
\newblock \href {https://doi.org/10.18653/v1/P18-2010} {Unsupervised semantic
  frame induction using triclustering}.
\newblock In \emph{Proceedings of the 56th Annual Meeting of the Association
  for Computational Linguistics (Volume 2: Short Papers)}, pages 55--62,
  Melbourne, Australia. Association for Computational Linguistics.

\bibitem[{Warstadt et~al.(2020)Warstadt, Parrish, Liu, Mohananey, Peng, Wang,
  and Bowman}]{warstadt-etal-2020-blimp-benchmark}
Alex Warstadt, Alicia Parrish, Haokun Liu, Anhad Mohananey, Wei Peng, Sheng-Fu
  Wang, and Samuel~R. Bowman. 2020.
\newblock \href {https://doi.org/10.1162/tacl_a_00321} {{BL}i{MP}: The
  benchmark of linguistic minimal pairs for {E}nglish}.
\newblock \emph{Transactions of the Association for Computational Linguistics},
  8:377--392.

\bibitem[{Wei et~al.(2022)Wei, Wang, Schuurmans, Bosma, Chi, Le, and
  Zhou}]{wei-etal-2022-chain}
Jason Wei, Xuezhi Wang, Dale Schuurmans, Maarten Bosma, Ed~Chi, Quoc Le, and
  Denny Zhou. 2022.
\newblock Chain of thought prompting elicits reasoning in large language
  models.
\newblock \emph{arXiv preprint arXiv:2201.11903}.

\bibitem[{West et~al.(2022)West, Bhagavatula, Hessel, Hwang, Jiang, Le~Bras,
  Lu, Welleck, and Choi}]{west-etal-2022-symbolic}
Peter West, Chandra Bhagavatula, Jack Hessel, Jena Hwang, Liwei Jiang, Ronan
  Le~Bras, Ximing Lu, Sean Welleck, and Yejin Choi. 2022.
\newblock \href {https://doi.org/10.18653/v1/2022.naacl-main.341} {Symbolic
  knowledge distillation: from general language models to commonsense models}.
\newblock In \emph{Proceedings of the 2022 Conference of the North American
  Chapter of the Association for Computational Linguistics: Human Language
  Technologies}, pages 4602--4625, Seattle, United States. Association for
  Computational Linguistics.

\bibitem[{Wolf et~al.(2020)Wolf, Debut, Sanh, Chaumond, Delangue, Moi, Cistac,
  Rault, Louf, Funtowicz, Davison, Shleifer, von Platen, Ma, Jernite, Plu, Xu,
  Le~Scao, Gugger, Drame, Lhoest, and Rush}]{wolf-etal-2020-transformers}
Thomas Wolf, Lysandre Debut, Victor Sanh, Julien Chaumond, Clement Delangue,
  Anthony Moi, Pierric Cistac, Tim Rault, Remi Louf, Morgan Funtowicz, Joe
  Davison, Sam Shleifer, Patrick von Platen, Clara Ma, Yacine Jernite, Julien
  Plu, Canwen Xu, Teven Le~Scao, Sylvain Gugger, Mariama Drame, Quentin Lhoest,
  and Alexander Rush. 2020.
\newblock \href {https://doi.org/10.18653/v1/2020.emnlp-demos.6} {Transformers:
  State-of-the-art natural language processing}.
\newblock In \emph{Proceedings of the 2020 Conference on Empirical Methods in
  Natural Language Processing: System Demonstrations}, pages 38--45, Online.
  Association for Computational Linguistics.

\bibitem[{Yuan et~al.(2018)Yuan, Ren, He, Zhang, Geng, Huang, Ji, Lin, and
  Han}]{yuan-etal-2018-open}
Quan Yuan, Xiang Ren, Wenqi He, Chao Zhang, Xinhe Geng, Lifu Huang, Heng Ji,
  Chin-Yew Lin, and Jiawei Han. 2018.
\newblock Open-schema event profiling for massive news corpora.
\newblock In \emph{Proceedings of the 27th ACM International Conference on
  Information and Knowledge Management}.

\bibitem[{Zhang et~al.(2020)Zhang, Khashabi, Song, and
  Roth}]{zhang-etal-2020-transomcs}
Hongming Zhang, Daniel Khashabi, Yangqiu Song, and Dan Roth. 2020.
\newblock Transomcs: From linguistic graphs to commonsense knowledge.
\newblock In \emph{Proceedings of International Joint Conference on Artificial
  Intelligence (IJCAI) 2020}.

\bibitem[{Zhao et~al.(2022)Zhao, Hessel, Yu, Lu, Zellers, and
  Choi}]{zhao-etal-2022-connecting}
Yanpeng Zhao, Jack Hessel, Youngjae Yu, Ximing Lu, Rowan Zellers, and Yejin
  Choi. 2022.
\newblock \href {https://doi.org/10.18653/v1/2022.naacl-main.333} {Connecting
  the dots between audio and text without parallel data through visual
  knowledge transfer}.
\newblock In \emph{Proceedings of the 2022 Conference of the North American
  Chapter of the Association for Computational Linguistics: Human Language
  Technologies}, pages 4492--4507, Seattle, United States. Association for
  Computational Linguistics.

\bibitem[{Zheng et~al.(2019)Zheng, Cao, Xu, and
  Bian}]{zheng-etal-2019-doc2edag}
Shun Zheng, Wei Cao, Wei Xu, and Jiang Bian. 2019.
\newblock \href {https://doi.org/10.18653/v1/D19-1032} {{D}oc2{EDAG}: An
  end-to-end document-level framework for {C}hinese financial event
  extraction}.
\newblock In \emph{Proceedings of the 2019 Conference on Empirical Methods in
  Natural Language Processing and the 9th International Joint Conference on
  Natural Language Processing (EMNLP-IJCNLP)}, pages 337--346, Hong Kong,
  China. Association for Computational Linguistics.

\bibitem[{Zhou et~al.(2020)Zhou, Zhang, Cui, and
  Huang}]{zhou-etal-2020-evaluating}
Xuhui Zhou, Yue Zhang, Leyang Cui, and Dandan Huang. 2020.
\newblock Evaluating commonsense in pre-trained language models.
\newblock In \emph{Proceedings of the AAAI Conference on Artificial
  Intelligence}.

\end{thebibliography}
\bibliographystyle{acl_natbib}

\clearpage

\appendix

\section{Running Environments and Hyper-parameters}
\label{sec:running_environments_and_hyper_parameters}

We run all experiments on a single server with 56 CPU cores and an Nvidia TITAN RTX GPU, except the BLOOM model runs on a cluster with 5 NVIDIA A100 GPUs.
Our codes rely on the Huggingface Library~\citep{wolf-etal-2020-transformers}.
For all corpora, we filter sentences that are too long or contain too many numerical tokens.
For baselines, we dump Wikipedia 20220301 version\footnote{https://dumps.wikimedia.org} and pre-process them by WikiExtractor~\citep{giusepppe-2015-wikiextractor} as background corpora.
And we use the OntoNotes sense grouping\footnote{http://verbs.colorado.edu/html\_groupings} as input verb sense dictionary for all clustering baselines.
Table~\ref{tab:shared_hyper_parameters} and Table~\ref{tab:specific_hyper_parameters} show the detailed hyper-parameters.

\begin{table}[!h]
\centering
\resizebox{\columnwidth}{!}{
\begin{tabular}{ccc}  
\hline 
\textbf{Shared Hyper-parameter} & \textbf{ESHer} & \textbf{Baselines} \\
\hline
Max Number of Tokens & \multicolumn{2}{c}{256} \\
Ratio of Numerical Tokens & \multicolumn{2}{c}{0.25} \\
Min Frequency of Verbs & \multicolumn{2}{c}{3} \\
Salient Ratio of Verbs & \multicolumn{2}{c}{0.25} \\
Min Frequency of Arguments & \multicolumn{2}{c}{3} \\
Salient Ratio of Arguments & \multicolumn{2}{c}{0.25} \\
Random Seed & \multicolumn{2}{c}{1234} \\
\hline 
\end{tabular}
}
\caption{Shared hyper-parameters for ESHer and other baselines in Section~\ref{sec:experiments}.}
\label{tab:shared_hyper_parameters}
\end{table}

\begin{table}[!h]
\centering
\resizebox{\columnwidth}{!}{ 
\begin{tabular}{ccc}  
\hline 
\textbf{Specific Hyper-parameter} & \textbf{English} & \textbf{Chinese} \\
\hline 
& \multicolumn{2}{c}{\textbf{ESHer}} \\
In-context Demonstrations $m$ & 8 & 9 \\
Candidates $n$ & \multicolumn{2}{c}{3} \\
PageRank $\beta$& \multicolumn{2}{c}{0.8} \\
PageRank $T$ & \multicolumn{2}{c}{300} \\
PageRank $\epsilon$ & \multicolumn{2}{c}{1e-6} \\
$\lambda_1$ & \multicolumn{2}{c}{1} \\
$\lambda_2$ & \multicolumn{2}{c}{1} \\
$\lambda_3$ & \multicolumn{2}{c}{3} \\
$\lambda_4$ & \multicolumn{2}{c}{1} \\
$\lambda_5$ & \multicolumn{2}{c}{1} \\
Threshold of Confidence & \multicolumn{2}{c}{1/3} \\
\hline 
& \multicolumn{2}{c}{\textbf{JCSC}~\citeyearpar{huang-etal-2016-liberal}} \\
\# Iteration Epoch & \multicolumn{2}{c}{100} \\
\hline
& \multicolumn{2}{c}{\textbf{Triframes}~\citeyearpar{ustalov-etal-2018-unsupervised-semantic}} \\
\# Neighbor & \multicolumn{2}{c}{10} \\
Weight & \multicolumn{2}{c}{0} \\
\hline
& \multicolumn{2}{c}{\textbf{ETypeClus}~\citeyearpar{shen-etal-2021-corpus}} \\
Aggregation Method & \multicolumn{2}{c}{Concat}  \\
Batch Size & \multicolumn{2}{c}{64} \\
Distribution & \multicolumn{2}{c}{Softmax} \\
Gamma & \multicolumn{2}{c}{0.02} \\
Hidden Dimension & \multicolumn{2}{c}{[500, 500, 1000, 100]} \\
Learning Rate & \multicolumn{2}{c}{0.001} \\
Pre-train Epoch & \multicolumn{2}{c}{1000} \\
Separately Decode & \multicolumn{2}{c}{False} \\
Sort Method & \multicolumn{2}{c}{Discriminative} \\
Temperature & \multicolumn{2}{c}{0.1} \\
Threshold & \multicolumn{2}{c}{0.05} \\
Train Epoch & \multicolumn{2}{c}{100} \\
Update Interval & \multicolumn{2}{c}{100} \\
Use Frequency & \multicolumn{2}{c}{False} \\
\hline
\end{tabular}
}
\caption{Specific hyper-parameters for ESHer and other baselines in Section~\ref{sec:experiments}..}
\label{tab:specific_hyper_parameters}
\end{table}

\begin{figure*}[!t]
\centering
\includegraphics[width=\textwidth]{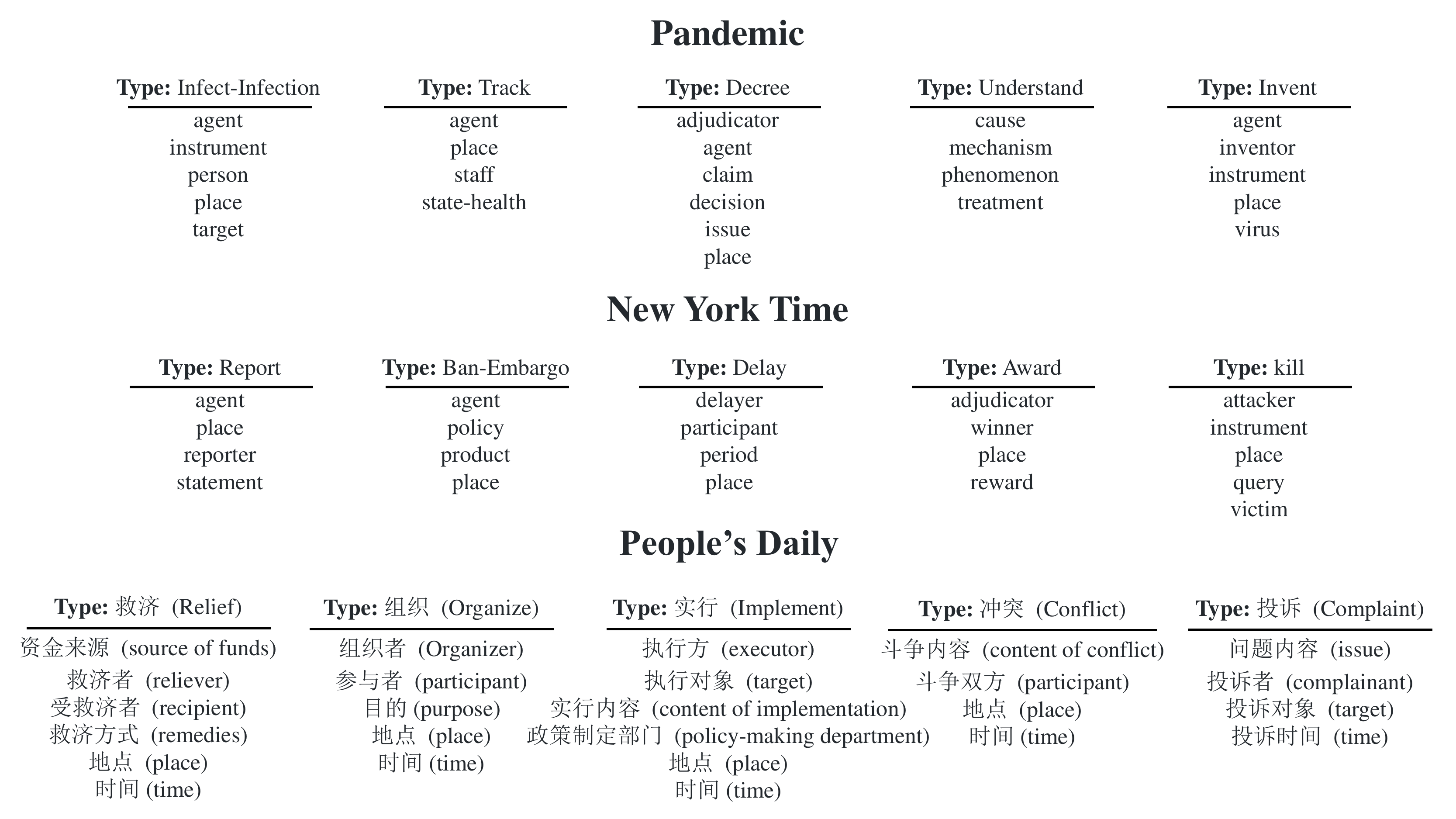}
\caption{
Schemas induced on varying domains.
}
\label{fig:more}
\end{figure*}

\section{Evaluation Metrics for Event Mention Clustering}
\label{sec:evaluation_metrics_for_event_mention_clustering}

We implement all metrics based on the Scikit-learn codebase~\citep{pedregosa-etal-2011-scikit}.
Following~\citet{shen-etal-2021-corpus}, we denote the ground truth clusters as $C^{*}$, the predicted clusters as $C$, and the total number of event mentions as $N$.

\begin{itemize}
\item \textbf{ARI}~\citep{hubert-etal-1985-comparing} measures the similarity between two cluster assignments based on the number of pairs in the same/different clusters.
Let $TP(T N)$ denotes the number of element pairs in the same (different) cluster(s) in both $C^{*}$ and $C$.
Then, ARI is calculated as follows:
\begin{equation}
\begin{aligned}
ARI &= \frac{RI - \mathbb{E}(EI)}{\max RI - \mathbb{E}(EI)} \\
RI &= \frac{TP + TN}{N}
\end{aligned}
\end{equation}
where $\mathbb{E}(EI)$ is the expected $RI$ of random assignments.
\item \textbf{NMI} denotes the normalized mutual information between two cluster assignments. 
Let $MI(\cdot; \cdot)$ be the Mutual Information between two cluster assignments, and $H(\cdot)$ denotes the Entropy. 
Then the NMI is formulated as follows:
\begin{equation}
\begin{aligned}
NMI = \frac{2 * MI(C^{*}; C)}{H(C^{*}) + H(C)}
\end{aligned}
\end{equation}
\item \textbf{BCubed-F1}~\citep{bagga-baldwin-1998-entity-based} estimates the quality of the generated cluster assignment by aggregating the precision and recall of each element. 
B-Cubed precision, recall, and F1 are thus calculated as follows:
\begin{equation}
\begin{aligned}
\emph{BCubed-P} &= \frac{1}{N} \sum_{i=1}^{N} \frac{|C(e_i) \cap C^{*}(e_i)|}{|C(e_i)|} \\
\emph{BCubed-R} &= \frac{1}{N} \sum_{i=1}^{N} \frac{|C(e_i) \cap C^{*}(e_i)|}{|C^{*}(e_i)|} \\
\emph{BCubed-F1} &= \frac{2}{\emph{BCubed-P}^{-1} + \emph{BCubed-P}^{-1}}
\end{aligned}
\end{equation}
where $C^{*}(\cdot)$ ($C(\cdot)$) is the mapping function from an element to its ground truth (predicted) cluster.
\end{itemize}

\section{More ESHer Outputs}
\label{sec:more_esher_outputs}
In this section, we provide more induced schema from varying domains, including pandemic (Cov-19 and Pandemic~\citep{shen-etal-2021-corpus}) and daily news (New York Time\footnote{https://catalog.ldc.upenn.edu/LDC2008T19} and People's Daily 1946-2001\footnote{http://en.people.cn}).

\end{document}